\documentclass{article}
\usepackage{spconf,amsmath,graphicx,multirow}
\usepackage{CJKutf8}

\title{Learn to Code-Switch: Data Augmentation using Copy Mechanism \\ on Language Modeling}
\name{Genta Indra Winata, Andrea Madotto, Chien-Sheng Wu, Pascale Fung\thanks{This work is partially funded by ITS/319/16FP of the Innovation Technology Commission, HKUST 16214415 \& 16248016 of Hong Kong Research Grants Council, and RDC 1718050-0 of EMOS.AI.}}
\address{
  Center for Artificial Intelligence Research (CAiRE)\\
  Department of Electronic and Computer Engineering\\
  Hong Kong University of Science and Technology, Clear Water Bay, Hong Kong\\ 
  \tt \{giwinata, amadotto, cwuak\}@connect.ust.hk, pascale@ece.ust.hk
}
\begin{document}
%
\maketitle
\begin{abstract}
Building large-scale datasets for training code-switching language models is challenging and very expensive. To alleviate this problem using parallel corpus has been a major workaround. However, existing solutions use linguistic constraints which may not capture the real data distribution. In this work, we propose a novel method for learning how to generate code-switching sentences from parallel corpora. Our model uses a Seq2Seq model in combination with pointer networks to align and choose words from the monolingual sentences and form a grammatical code-switching sentence. In our experiment, we show that by training a language model using the augmented sentences we improve the perplexity score by $10\%$ compared to the LSTM baseline. 
\end{abstract}
\begin{keywords}
code-switch, bilingual, copy attention, language modeling, data augmentation
\end{keywords}
\section{Introduction}
\label{sec:intro}
Language mixing has been a common phenomenon in multilingual communities. It is motivated in response to social factors as a way of communication in a multicultural society. From a sociolinguistic perspective, individuals do code-switching in order to construct an optimal interaction by accomplishing the conceptual, relational-interpersonal, and discourse-presentational meaning of conversation \cite{bhatt2011code}. In its practice, the variation of code-switching will vary due to the traditions, beliefs, and normative values in the respective communities. A number of studies \cite{poplack1978syntactic,pfaff1979constraints,poplack1980sometimes, belazi1994code} found that code-switching is not produced indiscriminately, but follows syntactic constraints. Many linguists formulated various constraints to define a general rule for code-switching \cite{poplack1978syntactic, poplack1980sometimes,belazi1994code}. However, the constraints are not enough to make a good generalization of real code-switching constraints, and they have not been tested in large-scale corpora for many language pairs.

One of the biggest problem in code-switching is collecting large scale corpora. Speech data have to be collected from a spontaneous speech by bilingual speakers and the code-switching has to be triggered naturally during the conversation. In order to solve the data scarcity issue, code-switching data generation is useful to increase the volume and variance. A linguistics constraint-driven generation approach such as equivalent constraint \cite{li2012code,pratapa2018language} is not restrictive to languages with distinctive grammar structure.
\begin{figure*}[!t]
  \centering
  \includegraphics[width=0.78\linewidth]{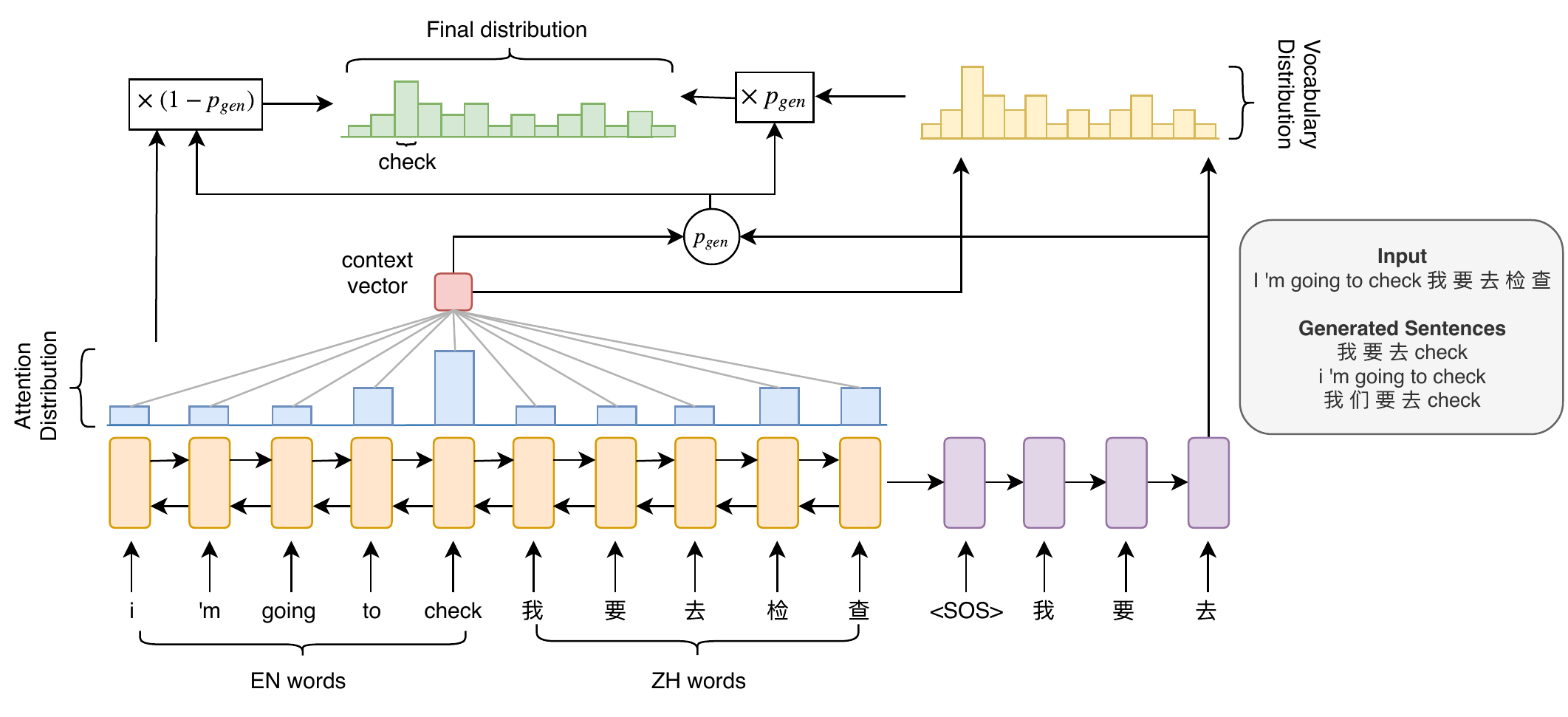}
  \caption{Pointer Generator Networks \cite{SeeP17-1099}. The figure  shows the example of input and 3-best generated sentences.}
  \label{fig:pointer-generation}
\end{figure*}

In this paper, we propose a novel language-agnostic method to learn how to generate code-switching sentences by using a pointer-generator network \cite{SeeP17-1099}. The model is trained from concatenated sequences of parallel sentences to generate code-switching sentences, constrained by code-switching texts. The pointer network copies words from both languages and pastes them into the output, generating code switching sentences in matrix language to embedded language and vice versa. The attention mechanism helps the decoder to generate meaningful and grammatical sentences without needing any sequence alignment. This idea is also in line with code-mixing by borrowing words from the embedded language \cite{lipski2005code} and intuitively, the copying mechanism can be seen as an end-to-end approach to translate, align, and reorder the given words into a grammatical code-switching sentence. This approach is the unification of all components in the work of \cite{li2012code} into a single computational model. A code-switching language model learned in this way is able to capture the patterns and constraints of the switches and mitigate the out-of-vocabulary (OOV) issue during sequence generation. 
~By adding the generated sentences and incorporating syntactic information to the training data, we achieve better performance by $10\%$ compared to an LSTM baseline \cite{winata2018code} and $5\%$ to the equivalent constraint.


\section{Related Work}
\label{sec:related-work}

The synthetic code-switching generation approach was introduced by adapting equivalence constraint on monolingual sentence pairs during the decoding step on an automatic speech recognition (ASR) model \cite{li2012code}. \cite{ying2014language} explored Functional Head Constraint, which was found to be more restrictive than the Equivalence Constraint, but complex to be implemented, by using a lattice parser with a weighted finite-state transducer. \cite{adel2013recurrent} extended the RNN by adding POS information to the input layer and factorized output layer with a language identifier. Then, Factorized RNN networks were combined with an n-gram backoff model using linear interpolation \cite{adel2013combination}. \cite{adel2015syntactic} added syntactic and semantic features to the Factorized RNN networks.~\cite{baheti2017curriculum} adapted an effective curriculum learning by training a network with monolingual corpora of both languages, and subsequently train on code-switched data. A further investigation of Equivalence Constraint and Curriculum Learning showed an improvement in language modeling \cite{pratapa2018language}. A multi-task learning approach was introduced to train the syntax representation of languages by constraining the language generator \cite{winata2018code}. 

A copy mechanism was proposed to copy words directly from the input to the output using an attention mechanism ~\cite{vinyals2015pointer}. This mechanism has proven to be effective in several NLP tasks including text summarization \cite{SeeP17-1099}, and dialog systems \cite{P18-1136}. The common characteristic of these tasks is parts of the output are exactly the same as the input source. For example, in dialog systems the responses most of the time have appeared in the previous dialog steps.

\section{Methodology}
\label{sec:methodology}
We use a sequence to sequence (Seq2Seq) model in combination with pointer and copy networks \cite{SeeP17-1099} to align and choose words from the monolingual sentences and generate a code-switching sentence. The models' input is the concatenation of the two monolingual sentences, denoted as $[w_1^{\ell_1},\dots,w_n^{\ell_1}, w_1^{\ell_2},\dots,w_m^{\ell_2} ]$, and the output is a code-switched sentence, denoted as $y_1^{cs},\dots,y_k^{cs}$. The main assumption is that almost all, the token present in the code-switching sentence are also present in the source monolingual sentences. Our model leverages this property by copying input tokens, instead of generating vocabulary words. This approach has two major advantages:~(1) the learning complexity decreases since it relies on copying instead of generating; (2) improvement in generalization, the copy mechanism could produce words from the input that are not present in the vocabulary.

\subsection{Pointer-generator Network}
Instead of generating words from a large vocabulary space using a Seq2Seq model with attention \cite{luong2015effective}, pointer-generator network~\cite{SeeP17-1099} is proposed to copy words from the input to the output using an attention mechanism and generate the output sequence using decoders. The network is depicted in Figure \ref{fig:pointer-generation}. For each decoder step, a generation probability $p_{gen}$ $\in$ [0,1] is calculated, which weights the probability of generating words from the vocabulary, and copying words from the source text.~$p_{gen}$ is a soft gating probability to decide whether generating the next token from the decoder or copying the word from the input instead. The attention distribution $a_t$ is a standard attention with general scoring \cite{luong2015effective}. It considers all encoder hidden states to derive the context vector. The vocabulary distribution $P_{vocab}(w)$ is calculated by concatenating the decoder state $s_t$ and the context vector $h_t^*$.
\begin{equation}
p_{gen} = \sigma (w_{h^*}^T h_t^* + w_s^T s_t + w_x^T x_t + b_{ptr})
\end{equation}
where $w_{h^*}, w_s, w_x$ are trainable parameters and $b_{ptr}$ is the scalar bias. The vocabulary distribution $P_{vocab}(w)$ and the attention distribution $a^t$ are weighted and summed to obtain the final distribution $P(w)$. The final distribution is calculated as follows:
\begin{equation}
P(w) = p_{gen} P_{vocab}(w) + (1 - p_{gen})\sum_{i:w_i=w}{a_i^t}
\end{equation}
We use a beam search to select $N$-best code-switching sentences and concatenate the generated sentence with the training set to form a larger dataset. The result of the generated code-switching sentences is showed in Table \ref{data-statistics-phase-2}. As our baseline, we compare our proposed method with three other models: (1) We use Seq2Seq with attention; (2) We generate sequences that satisfy Equivalence Constraint \cite{li2012code}. The constraint doesn't allow any switch within a crossing of two word alignments. We use FastAlign \cite{N13-1073} as the word aligner\footnote{The code can be found at https://github.com/clab/fast\_align}; (3) We also form sentences using the alignments without any constraint. The number of the generated sentences are equivalent to 3-best data from the pointer-generator model. To increase the generation variance, we randomly permute each alignment to form a new sequence.

\begin{table}[!t]
\centering
\small
\caption{Data Statistics of SEAME Phase II and Generated Sequences using Pointer-generator Network \cite{winata2018code}.}
\label{data-statistics-phase-2}
\begin{tabular}{|r|c|c|c|c|c|} 
\hline
\multicolumn{1}{|l|}{\multirow{2}{*}{}} & \multicolumn{3}{c|}{\textbf{SEAME Phase II}} & \multicolumn{2}{c|}{\textbf{Generated Seqs}}       \\ 
\cline{2-6}
\multicolumn{1}{|l|}{}                                            & \textbf{Train}  & \textbf{Dev}  & \textbf{Test}  & \textbf{1-best}  & \textbf{3-best}   \\ 
\hline
\# Speakers                                                       & 138             & 8             & 8              & -                & -                 \\ 
\hline
\# Utterances                                                     & 78,815          & 4,764         & 3,933          & 78,815           & 236,445           \\ 
\hline
\# Tokens                                                         & 1.2M            & 65K           & 60K            & -                & -                 \\ 
\hline
\begin{tabular}[c]{@{}r@{}}\# Tokens\\ Preprocessed \end{tabular} & 978K            & 53K           & 48K            & 945K             & 2.8M              \\ 
\hline
Avg. segment                                                      & 4.21            & 3.59          & 3.99           & 4.77             & 4.31              \\ 
\hline
Avg. switches                                                     & 2.94            & 3.12          & 3.07           & 2.51             & 2.79              \\
\hline
\end{tabular}
\end{table}

\subsection{Language Modeling}
The quality of the generated code-switching sentences is evaluated using a language modeling task. Indeed, if the perplexity in this task drops consistently we can assume that the generated sentences are well-formed. Hence, we use an LSTM language model with weight tying \cite{E17-2025} that can capture an unbounded number of context words to approximate the probability of the next word. 
Syntactic information such as Part-of-speech (POS) $[p_1, ..., p_T]$ is added to further improve the performance. The POS tags are generated phrase-wise using pretrained English and Chinese Stanford POS Tagger \cite{toutanova2003feature} by adding a word at a time in a unidirectional way to avoid any intervention from future information. The word and syntax unit are represented as a vector $x^w$ and $x^p$ respectively. Next, we concatenate both vectors and use it as an input $[x^w|x^p]$ to an LSTM layer similar to~\cite{winata2018code}.

\begin{table}[!t]
\centering
\small
\caption{Code-Switching Sentence Generation Results. Higher BLEU and lower perplexity (PPL) is better.}
\label{results-sentence-generation}
\begin{tabular}{|l|c|c|c|}
\hline
\multirow{2}{*}{} & \multicolumn{2}{c|}{\textbf{Dev}} & \multicolumn{1}{c|}{\textbf{Test}} \\ \cline{2-4} 
& \multicolumn{1}{c|}{\textbf{BLEU}} & \multicolumn{1}{c|}{\textbf{PPL}} & \multicolumn{1}{c|}{\textbf{BLEU}} \\ \hline
seq2seq with attention & 53.71 & 5.89 & 56.10 \\ \hline
pointer-generator & \textbf{55.19} & \textbf{4.61} & \textbf{59.68} \\ \hline
\end{tabular}
\end{table}

\begin{table}[!t]
\small
\centering
\caption{Language Modeling Results (in perplexity).}
\label{lm-results}
\begin{tabular}{|l|c|c|c|c|}
\hline
\multirow{2}{*}{} & \multicolumn{2}{c|}{\textbf{w/o syntax}} & \multicolumn{2}{c|}{\textbf{with syntax}} \\ \cline{2-5} & \textbf{Dev} & \textbf{Test} & \textbf{Dev} & \textbf{Test} \\ \hline
RNNLM \cite{winata2018code} & 178.35 & 171.27 & - & - \\
LSTM \cite{winata2018code} & 150.65 & 153.06 & 147.44 & 148.38 \\ \hline
\textbf{with augmentation} & & & & \\ 
random switch & 166.70 & 158.87 & 153.46 & 151.08 \\
equivalent-constraint & 149.72 & 147.03 & 147.48 & 145.05 \\
pointer-generator 1-best & 145.05 & 144.26 & 143.39 & 140.96              \\
pointer-generator 3-best & 144.69 & 143.84 & \textbf{142.84} & \textbf{138.91}     \\ \hline
\end{tabular}
\end{table}

\section{Experiment}
\subsection{Corpus}
In our experiment, we use a conversational
Mandarin-English code-switching speech corpus called SEAME Phase II (South East Asia Mandarin-English).~The data are collected from spontaneously spoken interviews and conversations in Singapore and Malaysia by bilinguals \cite{SEAME2015}.~As the data preprocessing, words are tokenized using Stanford NLP toolkit~\cite{manning-EtAl:2014:P14-5} and all hesitations and punctuations were removed except apostrophe.~The split of the dataset is identical to \cite{winata2018code} and it is showed in Table \ref{data-statistics-phase-2}.

\subsection{Training Setup}
In this section, we present the experimental settings for pointer-generator network and language model. Our experiment, our pointer-generator model has 500-dimensional hidden states and word embeddings. We use 50k words as our vocabulary for source and target. We evaluate our pointer-generator performance using BLEU\footnote{\textnormal{BLEU is computed using multi\_bleu.perl from MOSES package}} score. We take the best model as our generator and during the decoding stage, we generate 1-best and 3-best using beam search with a beam size of 5. For the input, we build a parallel monolingual corpus by translating the mixed language sequence using Google NMT\footnote{Google NMT Translate API}~to English ($w^{\ell_1}$) and Mandarin ($w^{\ell_2}$) sequences. Then, we concatenate the translated English and Mandarin sequences and assign code-switching sequences as the labels ($y^{cs}$). 

The baseline language model is trained using RNNLM \cite{mikolov2011rnnlm}\footnote{RNNLM toolkit from http://www.fit.vutbr.cz/$\mathtt{\sim}$imikolov/rnnlm/}. Then, we train our 2-layer LSTM models with a hidden size of 500 and unrolled for 35 steps. The embedding size is equal to the LSTM hidden size for weight tying. We optimize our model using SGD with initial learning rates of $\{10, 20\}$. If there is no improvement during the evaluation, we reduce the learning rate by a factor of 0.75. In each time step, we apply dropout to both embedding layer and recurrent network. The gradient is clipped to a maximum of 0.25. Perplexity measure is used in the evaluation.

\begin{figure}[!t]
  \centering
  \includegraphics[width=1.0\linewidth]{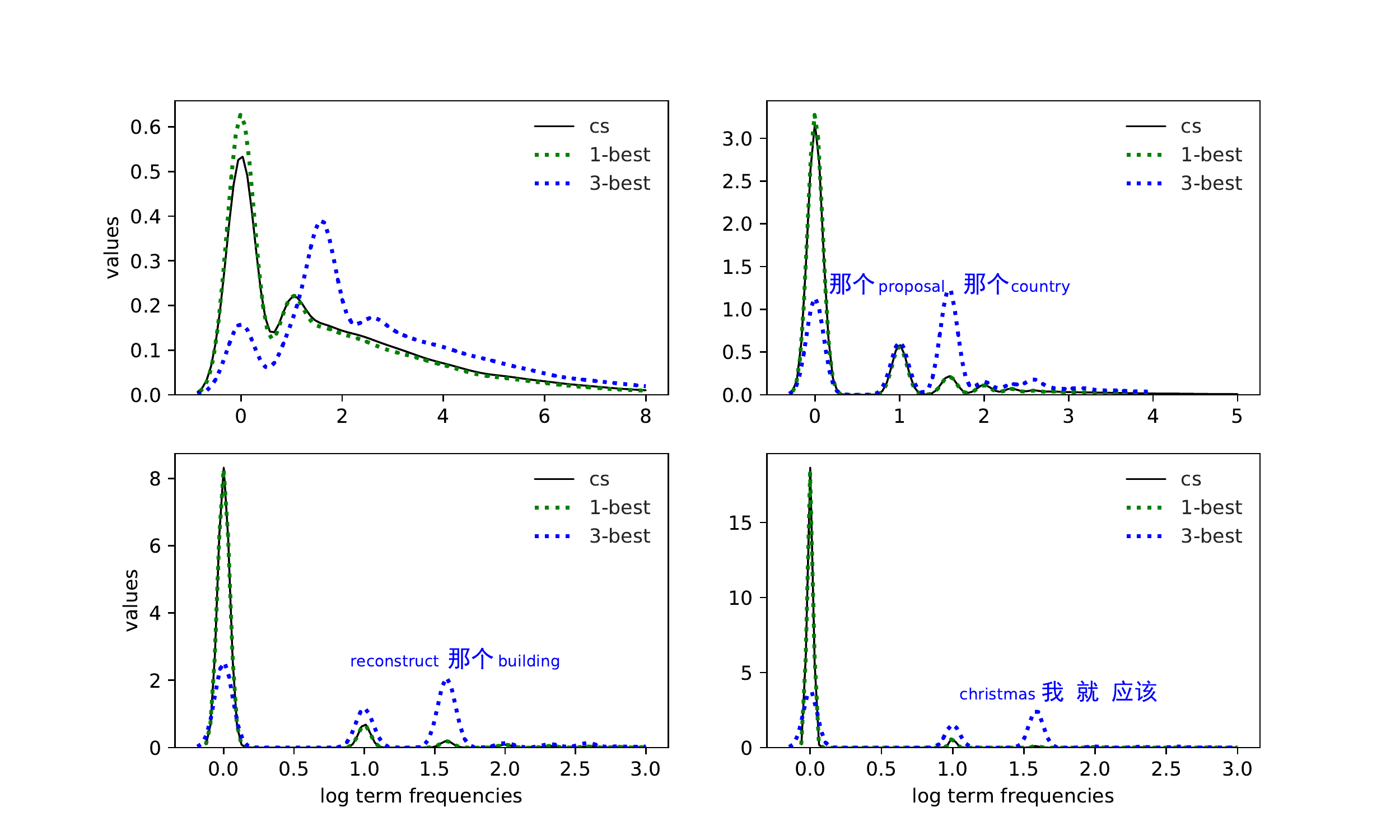}
  \caption{Univariate data distribution for unigram (top-left), bigram (top-right), trigram (bottom-left), and fourgram (bottom-right). The showed n-grams are sampled from 3-best data pointer-generator model.}
  \label{fig:dist-data}
\end{figure}

\section{Results}
\label{sec:results}
\begin{CJK*}{UTF8}{gbsn}
The pointer-generator significantly outperforms the Seq2Seq with attention model by 3.58 BLEU points on the test set as shown in Table \ref{results-sentence-generation}. Our language modeling result is given in Table \ref{lm-results}. Based on the empirical result, adding generated samples consistently improve the performance of all models with a moderate margin around 10\% in perplexity. After all, our proposed method still slightly outperforms the heuristic from linguistic constraint. In addition, we get a crucial gain on performance by adding syntax representation of the sequences.

\noindent\textbf{Change in data distribution: } To further analyze the generated result, we observed the distribution of real code-switching data and the generated code-switching data. From Figure \ref{fig:dist-data}, we can see that 1-best and real code-switching data have almost identical distributions. The distributions are left-skewed where the overall mean is less than the median. Interestingly, the distribution of the 3-best data is less skewed and generates a new set of n-grams such as ``那个\enspace(that) proposal" which was learned from other code-switching sequences. As a result, generating more samples effects the performance positively.

\noindent\textbf{Importance of Linguistic Constraint: } The result in Table \ref{lm-results} emphasizes that linguistic constraints have some significance in replicating the real code-switching patterns, specifically the equivalence constraint.~There is a slight reduction in perplexity around 6 points on the test set.~In addition, when we ignore the constraint, we lose performance because it still allows switches in the inversion grammar cases.

\noindent\textbf{Does the pointer-generator learn how to switch?} We found that our pointer-generator model generates sentences that have not been seen before.~The example in Figure \ref{fig:pointer-generation} shows that our model is able to construct a new well-formed sentence such as \textnormal{``我\enspace们\enspace要\enspace去\enspace (We want to) check"}. It is also shown that the pointer-generator model has the capability to learn the characteristics of the linguistic constraints from data without any word alignment between the matrix and embedded languages. On the other hand, training using 3-best data obtains better performance compared to 1-best data.~We found a positive correlation from Table \ref{data-statistics-phase-2}, where 3-best data is more similar to the test set in terms of segment length and number of switches compared to 1-best data. Adding more samples $N$ may improve the performance, but it will be saturated at a certain point. One way to solve this is by using more parallel samples.

\end{CJK*}

\section{Conclusion}
\label{sec:conclusion}
We introduce a new learning method for code-switching sentence generation using a parallel monolingual corpus that is applicable to any language pair. Our experimental result shows that adding generated sentences to the training data, effectively improves our model performance.~Combining the generated samples with code-switching dataset reduces perplexity. We get further performance gain after using syntactic information of the input. In future work, we plan to explore reinforcement learning for sequence generation and employ more parallel corpora.



\bibliographystyle{IEEEbib}
\bibliography{strings,refs}

\end{document}